\documentclass[preprint,12pt]{elsarticle}

\usepackage{amssymb}
\usepackage{lineno}

\journal{undisclosed journal}

\begin{document}

\begin{frontmatter}



\title{Tracked 3D Ultrasound and Deep Neural Network-based Thyroid Segmentation reduce Interobserver Variability in Thyroid Volumetry}


\author[nuk,rad]{Markus Krönke\corref{cor1}\fnref{first}}
\author[camp]{Christine Eilers\fnref{first}}
\author[camp]{Desislava Dimova}
\author[camp,med]{Melanie Köhler}
\author[nuk]{Gabriel Buschner}
\author[nuk]{Lilit Mirzojan}
\author[camp]{Lemonia Konstantinidou}
\author[rad]{Marcus R. Makowski}
\author[nuk,radboud]{James Nagarajah}
\author[camp,jhu]{Nassir Navab}
\author[nuk]{Wolfgang Weber}
\author[camp]{Thomas Wendler3}

\address[nuk]{Department of Nuclear Medicine, School of Medicine, Technical University of Munich, Munich, Germany}
\address[rad]{Department of Radiology, School of Medicine, Technical University of Munich, Munich, Germany}
\address[camp]{Chair for Computer Aided Medical Procedures and Augmented Reality, Department of Computer Science, Technical University of Munich, Garching near Munich, Germany}
\address[med]{Medical Faculty, Technical University of Munich, Munich, Germany}
\address[radboud]{Nuclear Medicine, Radboud University Medical Center, Nijmegen, the Netherlands}
\address[camp]{Chair for Computer Aided Medical Procedures, Whiting School of Engineering, Johns Hopkins University, Baltimore (MD), USA}

\cortext[cor1]{Corresponding author:\\Markus Krönke\\Klinikum rechts der Isar\\der Technischen Universität München\\Institut für diagnostische und interventionelle Radiologie\\Ismaninger Str. 22\\81675 München\\Tel:  +49 89 4140-8804\\Fax: +49 89 4140-7709\\markus.kroenke@tum.de}

\fntext[first]{These authors contributed equally to this work and are considered to be co-first authors.}

\begin{abstract}
\noindent \textbf{Background:} Thyroid volumetry is crucial in diagnosis, treatment and monitoring of thyroid diseases. However, conventional thyroid volumetry with 2D ultrasound is highly operator-dependent. This study compares 2D ultrasound and tracked 3D ultrasound with an automatic thyroid segmentation based on a deep neural network regarding inter- and intraobserver variability, time and accuracy. Volume reference was MRI.

\noindent \textbf{Methods:} 28 healthy volunteers (24 - 50 a) were scanned with 2D and 3D ultrasound as well as by MRI. Three physicians (MD 1, 2, 3) with different levels of experience (6, 4 and 1 a) performed three 2D ultrasound and three tracked 3D ultrasound scans on each volunteer. In the 2D scans the thyroid lobe volumes were calculated with the ellipsoid formula. A convolutional deep neural network (CNN) segmented the 3D thyroid lobes automatically. 26, 6, and 6 random lobe scans were used for training, validation and testing, respectively. On MRI (T1 VIBE sequence) the thyroid was manually segmented by an experienced medical doctor.

\noindent \textbf{Results:} MRI thyroid volumes ranged from $2.8$ to $16.7 ml$ (mean $7.4$, SD $3.05$). The CNN was trained to obtain an average dice score of $0.94$. The interobserver variability comparing two MDs showed mean differences for 2D and 3D respectively of $0.58$ to $0.52 ml$ (MD1 vs. 2), $-1.33$ to $-0.17 ml$ (MD1 vs. 3) and $-1.89$ to $-0.70 ml$ (MD2 vs. 3). Paired samples t-tests showed significant differences in two comparisons for 2D ($p=.140$, $p=.002$ and $p=.002$) and none for 3D ($p=.176$, $p=.722$ and $p=.057$). Intraobsever variability was similar for 2D and 3D ultrasound. Comparison of ultrasound volumes and MRI volumes by paired samples t-tests showed a significant difference for the 2D volumetry of all MDs ($p=.002$ , $p=.009$ , $p<.001$), and no significant difference for 3D ultrasound ($p=.292$, $p=.686$, $p=0.091$). Acquisition time was significantly shorter for 3D ultrasound.

\noindent \textbf{Conclusion:} Tracked 3D ultrasound combined with a CNN segmentation significantly reduces interobserver variability in thyroid volumetry and increases the accuracy of the measurements with shorter acquisition times.
\end{abstract}

\begin{keyword}
tracked 3D ultrasound \sep thyroid volumetry \sep ultrasound segmentation


\end{keyword}

\end{frontmatter}

%\linenumbers

\section{Introduction}
\label{sec:intro}

Exact thyroid volumetry plays an important role in monitoring and treatment of many thyroid diseases, e.g., the radioiodine therapy (RIT) in hyperthyroidism: The needed activity, commonly calculated using Marinelli's formula, is proportional to the thyroid mass~\cite{szumowski_calculation_2016}. The latter is obtained by using the thyroid volume measurements from two-dimensional ultrasound (2D US) using the ellipsoid formula and a correction factor~\cite{brunn_volumetric_1981}. Checkups are performed three and six months after RIT. If hyperthyroidism and a relevant thyroid volume are still detected, a second RIT is necessary. The probability of this second treatment is dependent on the dose concept, thyroid volume, and achieved organ dose~\cite{dietlein_radioiodtherapie_2016}. Therefore an underestimation of the measured thyroid volume could lead to an underestimated administered dose and thus to the need for a second RIT. On the other hand, an overdosage may lead to hypothyroidism and radiation-induced malignancies~\cite{mariani_role_2021}.

Currently, 2D US is the standard procedure for thyroid volumetry despite its high inter- and intraobserver variability. Three-dimensional (3D) US was first introduced in the 1970s, and it is believed to be superior to 2D US in terms of user dependency~\cite{lyshchik_three-dimensional_2004}. Similarly, the automatic compounding of 2D US sequences to generate 3D US has been shown to reduce variability~\cite{kojcev_reproducibility_2017}. However, the potential advantages of 3D US for volumetry only can be exploited clinically in combination with automatic segmentation. Only in this way 3D US can keep up the speed and ease of use of 2D US employing the ellipsoid formula. An automatic segmentation has for example been developed by Webb et al.~\cite{webb_automatic_2021}. They developed a network for the segmentation of nodules, thyroid, and cysts on US thyroid cineclips. Their work however focused on the network development and its accuracy with respect to manually segmented 2D ground truth. In contrast,compared  to our we present an analysis on volumetry and variability in a realistic setup using 3D compounded ultrasound volumes acquired with a tracked ultrasound system analysis. 

Analysis of intra- and interobserver variability in thyroid volumetry and the comparison of 2D and 3D US in this task is a fairly well analyzed topic in literature~\cite{lyshchik_three-dimensional_2004,kojcev_reproducibility_2017,reinartz_thyroid_2002,vulpoi_thyroid_2007,lee_intraobserver_2018,choi_inter-observer_2015,schlogl_novel_2006,lyshchik_accuracy_2004,malago_thyroid_2008,schlogl_use_2001,freesmeyer_multimodal_2014}. However, the combination of 3D US and machine learning for volumetry currently lacks exploration. 

In this paper, we therefore present a user dependency study on thyroid volumetry with 2D and 3D US in employing the first fully automatic segmentation of thyroid lobes by a deep neural network. Our aim is to analyze whether the hypothesis that 3D US imaging combined with this deep neural network is superior to conventional 2D US in terms of accuracy and variability of thyroid volume estimations is valid. Intra- and interobserver variability was calculated for both modalities and then compared with each other. The accuracy was estimated by comparing both US methods to Magnetic Resonance Imaging (MRI). 

In this paper our contributions are: 
\begin{itemize}
\item We introduce a first fully automatic method for 3D volumetry of thyroid lobes using tracked 3D US. 
\item We comprehensively evaluate our method in comparison to 2D US by analyzing intra- and interobserver variability and accuracy by comparing to MRI.
\item We evaluate the acquisition time performance of our method compared to 2D US.
\item We provide a neural network for thyroid segmentation from 3D US, that will become publicly available shortly.
\item We provide a novel tracked 3D US thyroid dataset as a first bigger and labelled open source dataset, that will become publicly available shortly.
\end{itemize}

\section{Materials and Methods}

\subsection{Materials}

This study was conducted in the scope of an Institutional Review Board Approval from the Ethical Commission of the Technical University of Munich (approved on April 1st, 2020; reference number 244/19 S). All volunteers gave written informed consent to the US and MRI scans.

\begin{figure}[h]
	\centering\includegraphics[width=0.9\linewidth]{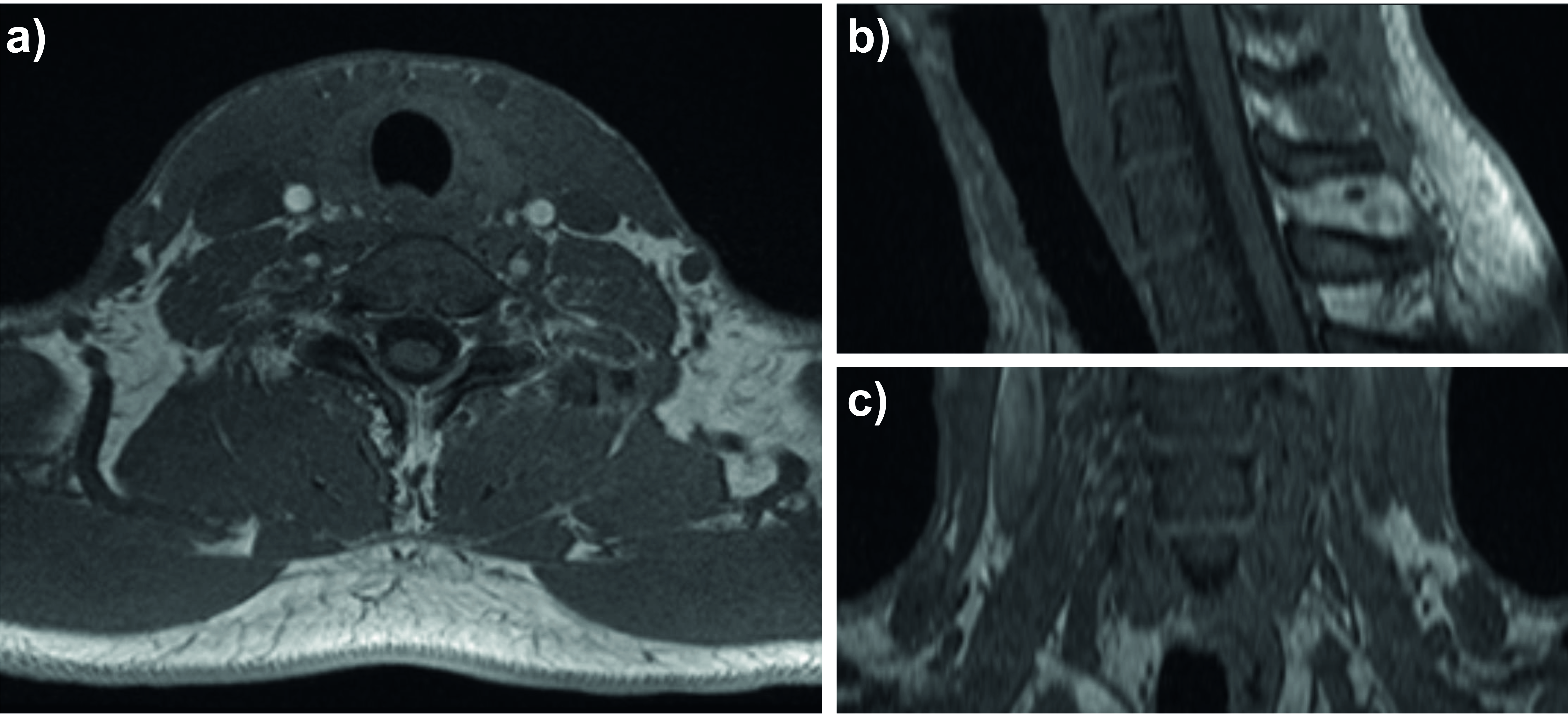}
	\caption{T1 VIBE MRI of the neck region for an exemplary volunteer, a) axial view, b) sagittal view, c) coronal view.}
	\label{fig:1}
\end{figure}

\begin{figure}[h]
	\centering\includegraphics[width=0.9\linewidth]{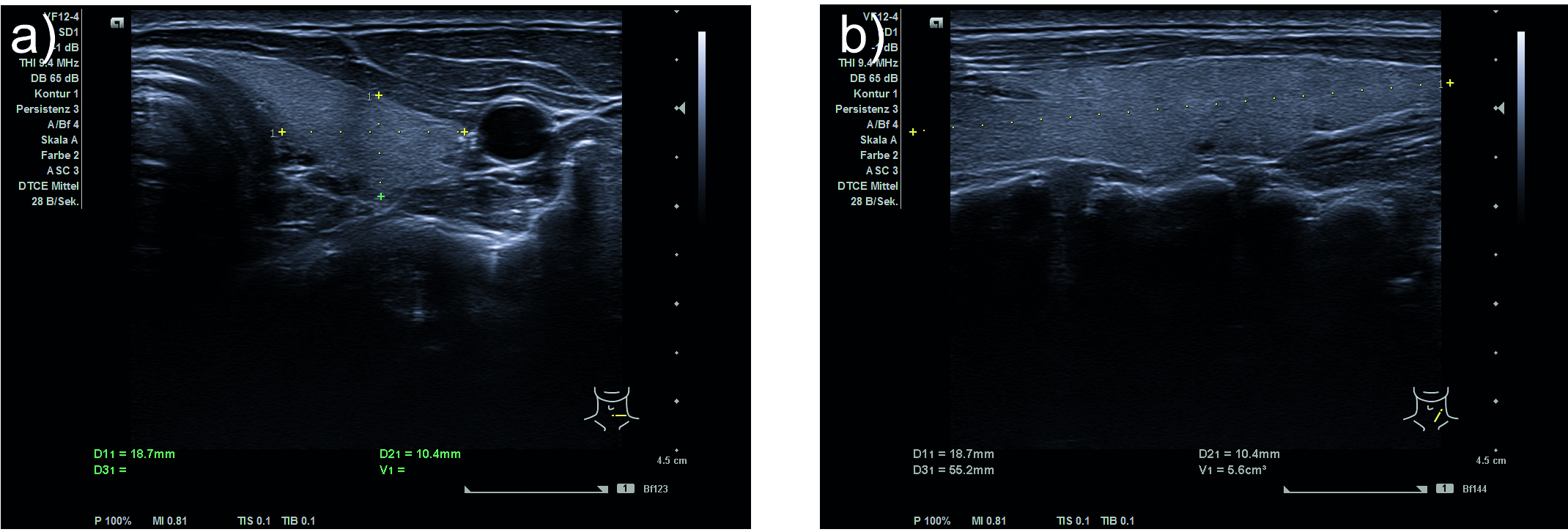}
	\caption{2D US scans of the neck region for an exemplary volunteer. a) Width and depth of right thyroid lobe marked, b) Length of right thyroid lobe marked.It can be seen that not the whole thyroid lobe is visible. This is a common issue for large thyroids.}
	\label{fig:2}
\end{figure}

Eligible participants had to be anamnestically healthy and at least 18 years old and were recruited from the hospital and research lab environment. Exclusion criteria for possible candidates were subtotal or total thyroidectomy, distinct postoperative or inherent changes in the neck anatomy and MRI contraindications (i.e. claustrophobia, neurostimulators, pacemaker, cochlear implants, medication pumps and shrapnels). 28 healthy volunteers, aged 23 to 50, 18 males and 10 females, participated in the study. . 
One high resolution T1 VIBE (Volumetric interpolated breath-hold examination) MRI scan of the neck was acquired with a Biograph mMR machine (Siemens Healthineers AG, Erlangen, Germany) with a field strength of 3 T. The resolution of the MRI image was $0.625\times0.625\times1.00 mm^3$ and the field of view $320\times320\times80 mm^3$ (Figure~\ref{fig:1}). The accuracy of the MRI segmentation was evaluated with a thyroid phantom (thyroid ultrasound training phantom, model 074, CIRS). According to the manufacturer, the thyroid volume of each phantom has slight variations of $\pm3$ to $5 ml$ but is close to $30 ml$. From the MRI scan segmentation we retrieved a volume of $26.33 ml$.
US images were obtained with an ACUSON NX3 machine (Siemens Healthineers AG, Erlangen, Germany) using a linear $12 MHz$ VF12-4 probe (Fig.~\ref{fig:2}). 

For the 3D US, electromagnetic tracking and video acquisition, a PIUR tUS system was used (piur imaging GmbH, Vienna, Austria). This device employs two 6D tracking sensors, operating at a frequency of $80 Hz$ with an accuracy in position of $1.40 mm$ RMS and in angle of $0.50 \deg$ RMS. Video sequences were acquired with $89 fps$. 3D US probes, which consist of matrix arrays, can also be used to acquire US volumes. However, these probes are not widely available, rather big and often do not cover the entire thyroid making a compounding necessary. 

The image processing (3D US compounding using image data and tracking information) and thyroid segmentation was done using ImFusion Suite Version 2.9.7 (ImFusion GmbH, Munich, Germany). PyTorch was used for the implementation of a deep neural network for automatic 3D US thyroid segmentation and Tensorflow was used for the evaluation of the network. The segmentation pipeline was implemented and trained by our team and differs from the one offered by  piur imaging in their latest release.

\subsection{General approach}
\begin{figure}[h]
	\centering\includegraphics[width=0.9\linewidth]{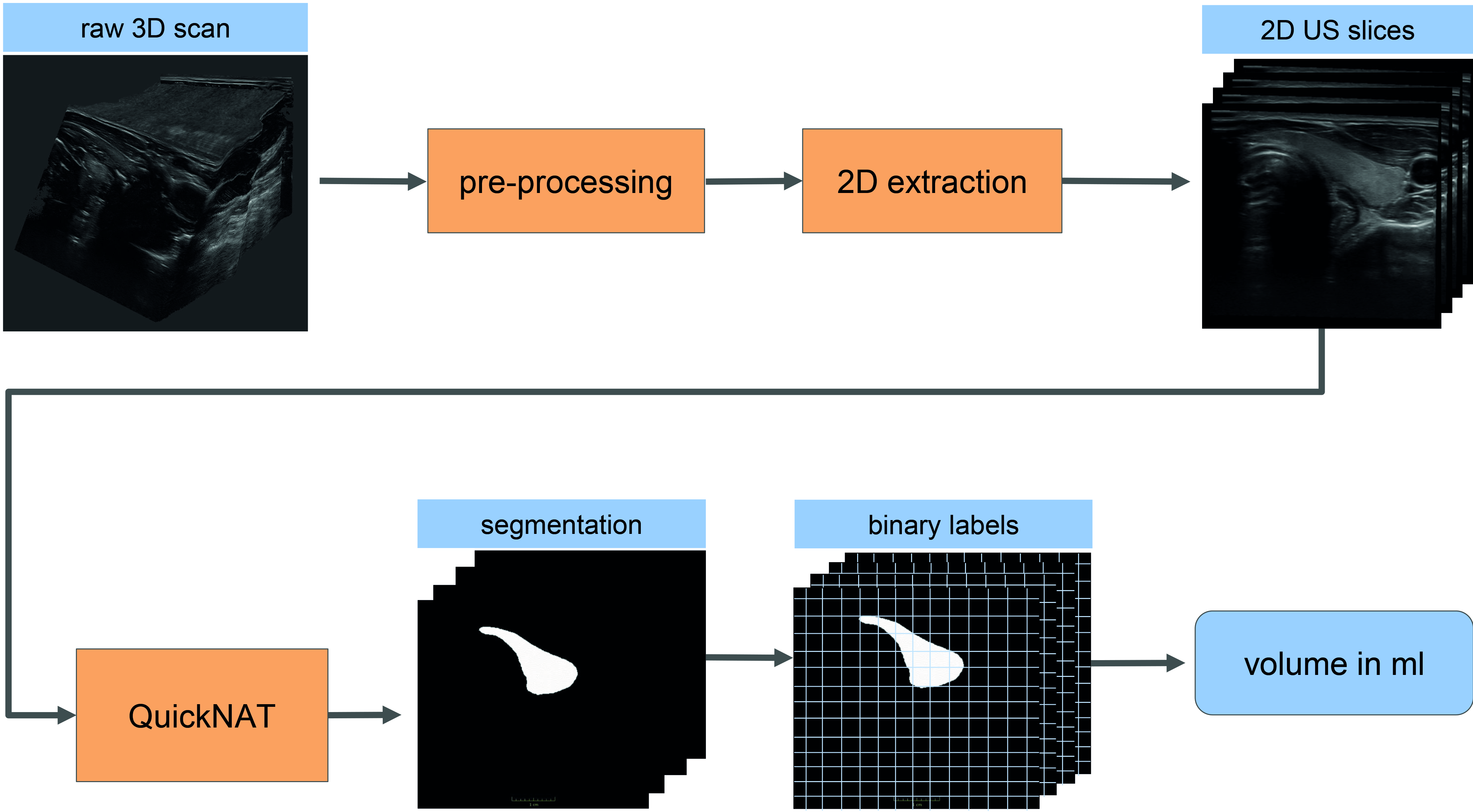}
	\caption{Schematic representation of the automatic segmentation workflow. The raw 3D scans of each lobe are pre-processed by rotation into axial view and resizing as well as centering. Then 2D images are extracted which serve as input to the network. The output of the network is a segmentation from which the volume is calculated by multiplication of pixels which include a segmentation with the pixel volume.}
	\label{fig:4}
\end{figure}

Three 2D and three 3D US scans for each volunteer were acquired by three physicians (MD1, MD2 and MD3) with different levels of experience (6, 4 and 1 years). The ellipsoid formula (correction factor $0.48$) was applied to estimate the thyroid volumes from the 2D US scans. No isthmus correction was included. To acquire a volumetric image for a 3D US, the physician slowly sweeps the transducer over both lobes (one sweep per lobe). These sweeps are automatically compounded into a 3D US stack. The 3D US thyroid volumes were calculated based on the automatic segmentation. The thyroid volumes, derived after manual segmentation of the  MRI scans by an experienced MD (8 years, cross-sectional imaging), were used as reference. The overall workflow is depicted in Figure~\ref{fig:4}.

\subsection{Deep neural network for automatic segmentation}
\begin{figure}[h]
	\centering\includegraphics[width=0.9\linewidth]{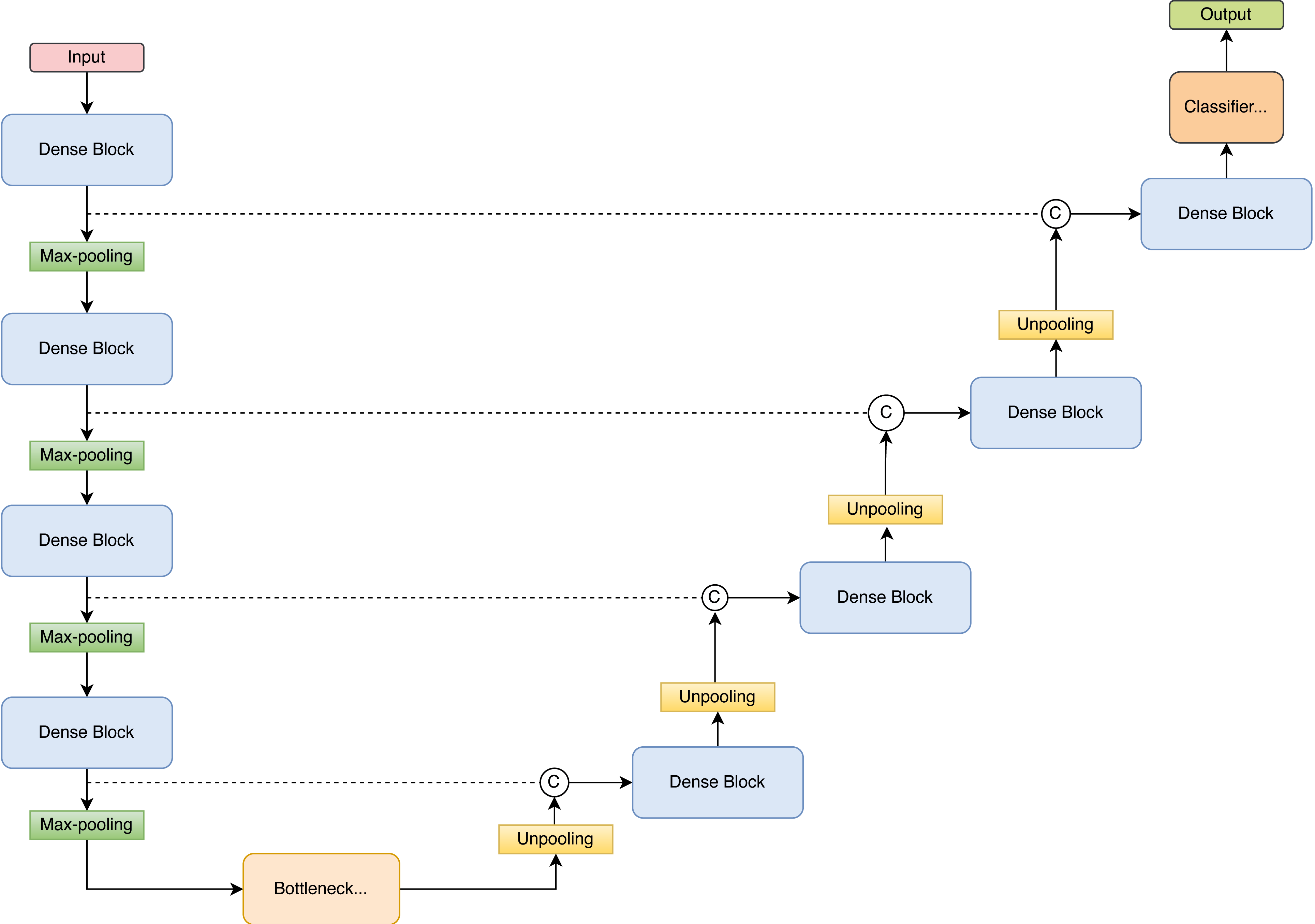}
	\caption{Illustration of the QuickNAT Architecture used here for automatically segmenting thyroid in 3D US volumes.}
	\label{fig:3}
\end{figure}

For the automatic segmentation of the 3D US scans, a fully convolutional deep neural network developed by Roy et al.~\cite{guha_roy_quicknat_2019} named QuickNAT was used (Fig.~\ref{fig:3}). The structure is similar to the U-net~\cite{ronneberger_u-net_2015}, although it can work with smaller datasets as well. 

QuickNAT consists of dense blocks of four encoders and four decoders with a bottleneck layer in between, as depicted in Fig.~\ref{fig:3}. Each dense block consists of three convolutional layers, two of $5\times5$ and one of $1\times1$ kernel size, preceded by a batch normalization layer and a Rectifier Linear Unit (ReLU). The bottleneck block of a $5\times5$ convolutional layer, including a batch normalization layer, restricts information flow in between the encoder and decoder. There are skip connections between the corresponding encoder and decoder blocks, to provide encoder feature information to the decoders directly and improve training through a path of gradient flow to the deeper layers. Each encoder is followed by a $2\times2$ max-pooling block, which reduces the spatial dimension of the feature maps. To achieve better segmentation for even smaller structures, there are un-pooling layers between the decoder blocks as the indices corresponding to the maximum activations are passed to the decoders. The final layer is a classifier block with softmax, which is a convolutional layer of $1\times1$ kernel size. It maps the input to a feature map with a given number of channels, two in this case, thyroid and background. Subsequently, the softmax layer transforms the channels into probability maps for each class. The loss function of the network is a combination of dice and cross-entropy losses.

26 , 6, and 6 random lobe scans were used for the training of the network, the validation and testing, respectively. The thyroid in these 3D US scans was manually segmented, slice by slice, by an experienced MD (8 years, cross-sectional imaging). 
Both lobes of the same volunteer were segmented manually which provided a total of $17.664$ slices for training (i.e. in average $226$ slices per lobe). This total was split into a training set of $14.352$ (13 patients) and a validation set of $3.312$ images (3 patients). The original architecture of QuickNAT was kept unchanged. A dropout of $0.5$ was applied to prevent overfitting. The weights on the dice and cross-entropy losses were kept equal. Edge weights were applied to improve the contour of the segmentation.

\begin{figure}[h]
	\centering\includegraphics[width=0.9\linewidth]{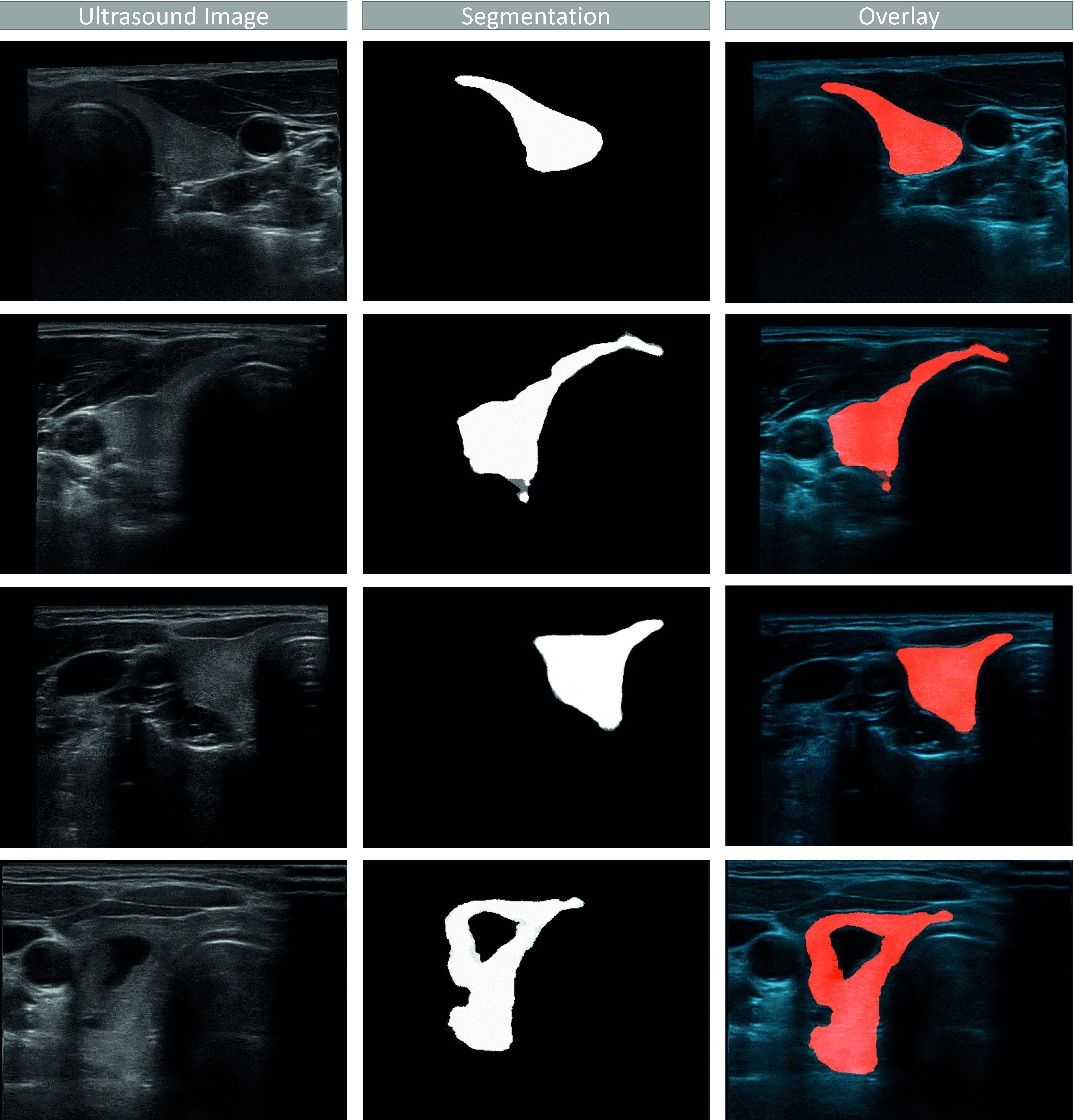}
	\caption{Ultrasound image (left), segmentation (middle) and overlay (right) of a good segmentation (first row), a segmentation which segmented too many voxels of the isthmus (second row), a segmentation with too few voxels of the isthmus segmented (third row) and a segmentation with a segmented nodule (fourth row).}
	\label{fig:5}
\end{figure}

The network was trained for 20 epochs with a batch size of 4 at a learning rate of $10^{-5}$. The dice score amounted to $0.95$, $0.94$ and $0.83$ for training, validation and test sets. The model was applied on the remaining data (13 volunteers) and the segmentations were used for the volume evaluations (Fig.~\ref{fig:5}).

\subsection{Statistics and Evaluation}

From the complete data set, scans from 16 volunteers were used for training and validation. Thus, the network was trained on images which were segmented by the same network in a later step for the volumetry estimation. We therefore split the evaluation into two groups, one including scans that were fed to the network during training (15 volunteers) (Group 1) and the other with only unseen data (13 volunteers) (Group 2) to be able to detect a potential bias.

To estimate the intraobserver variability of the acquisitions, we used a formula from Lyshchik et al.~\cite{lyshchik_three-dimensional_2004}.  A t-test was performed with 2D and 3D mean values to decide on whether both samples are significantly different. 
Additionally, we estimated the intraobserver variability as originally proposed by Choi et al~\cite{choi_inter-observer_2015}. To examine the intraobserver variability, we examined the differences between two volumes of the same measurement sequence on a Bland-Altman plot~\cite{bland_statistical_1986}. We show the mean as well as the $95\%$ limit of agreement ($\pm1.96 \times SD$). 

The interobserver variability was estimated by applying the method proposed by Choi et al.~\cite{choi_inter-observer_2015} as well. Here, the differences between volume measurements for each combination of two physicians were plotted in a Bland-Altman plot (Fig.~\ref{fig:7}).

Further, we applied paired samples t-tests to evaluate the difference between the mean volume measurements for each physician combination of two. In our case the objective was to determine whether the mean difference between the different measurements of the same volunteer is zero. We chose a significance level of $\alpha=0.05$ for all tests in both groups. 
To compare 2D and 3D US volume measurements to MRI, we applied paired samples t-tests between MRI the first scan sequence from each MD with the same significance level resulting in the same critical values per group.

\section{Results}

We will present all results for Group 1 and Group 2 separately. 

\subsection{Intraobserver variability}

\begin{figure}[h]
	\centering\includegraphics[width=0.9\linewidth]{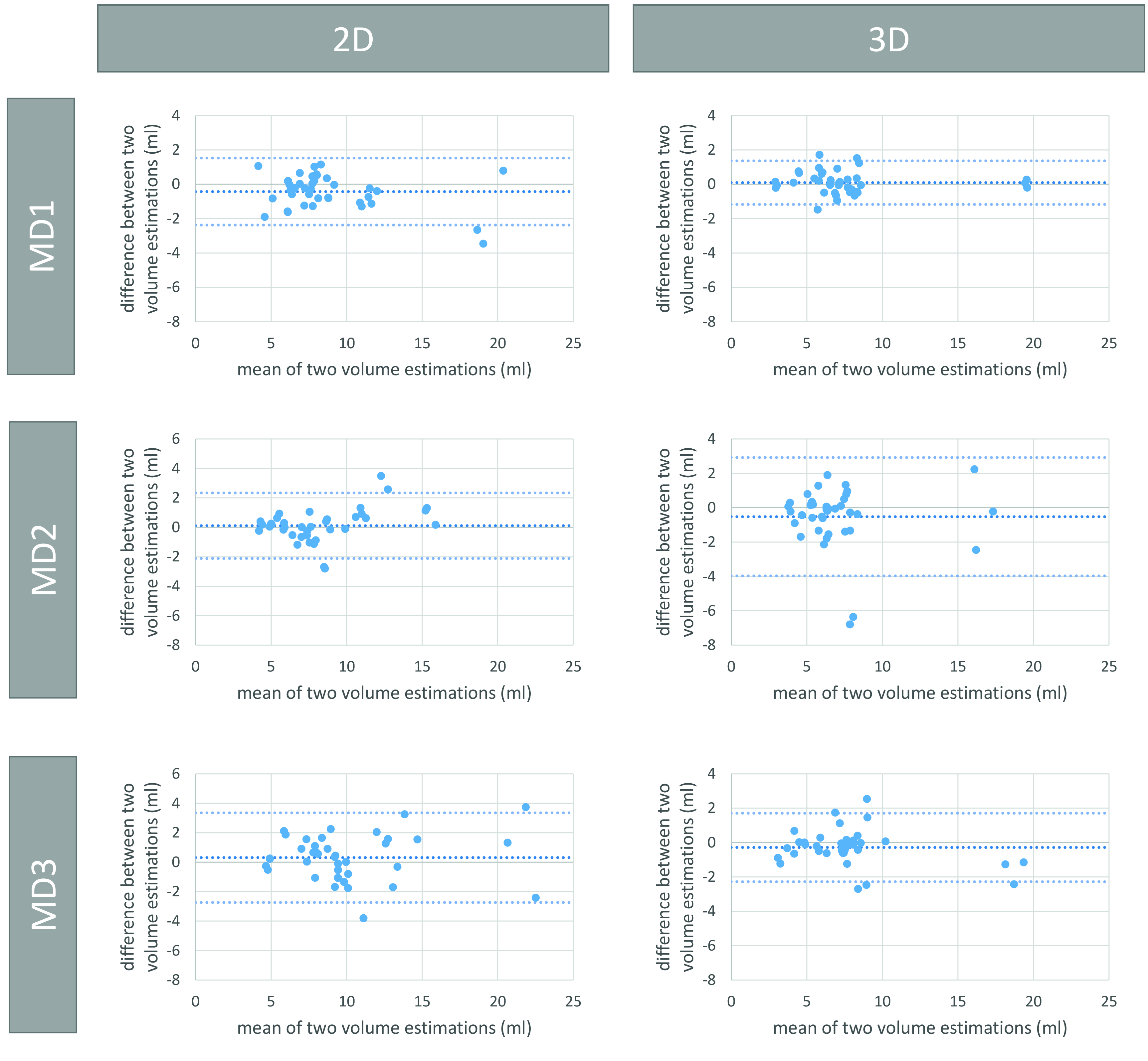}
	\caption{Bland-Altman plots for the intraobserver variability of each medical doctor from Group 2. All values are given in $ml$. Row 1, 2 and 3 show results for MD 1, 2, and 3, respectively. The columns show the results for 2D US (left) and 3D US measurements (right).}
	\label{fig:6}
\end{figure}

Intraobserver variability is summarized in Table~\ref{tab:1}. Significant differences were found comparing the sets from MD2 and MD3 in both Groups 1 and 2. Fig.~\ref{fig:6} depicts the Bland-Altman plots for the intraobserver variability of each MD with the 2D data on the left and the 3D data on the right for Group 2.

\begin{table}[h]
	\centering
	\begin{tabular}{c c c c c c c c}
		\hline
		MD & Exp. & \multicolumn{2}{c}{2D US} & \multicolumn{2}{c}{3D US} & \multicolumn{2}{c}{p-value}\\
		& (a) & Gr. 1 & Gr. 2 & Gr. 1 & Gr. 2 & Gr. 1 & Gr. 2 \\
		\hline
		1 & 6 & $12\pm5$ & $14\pm10$ & $12\pm10$ & $11\pm8$ & $.811$ & $.081$ \\
		2 & 4 & $18\pm10$ & $13\pm9$ & $15\pm10$ & $24\pm25$ & \boldmath $.003$ & \boldmath $.040$ \\
		3 & 1 & $15\pm10$ & $19\pm10$ & $8\pm8$ & $15\pm12$ & \boldmath $<.001$ & \boldmath $<.001$ \\
		\hline
	\end{tabular}
	\caption{Mean and standard deviation (mean$\pm$SD) in percent (\%) of the intraobserver variability for each MD in 2D and 3D for both groups (Gr.~1 and 2). The last column shows whether a significant difference exists between the averaged 2D and 3D volume estimation sets, being the significant different p-values in \textbf{bold}.}
	\label{tab:1}
\end{table}

\subsection{Interobserver variability}

\begin{figure}[h]
	\centering\includegraphics[width=0.9\linewidth]{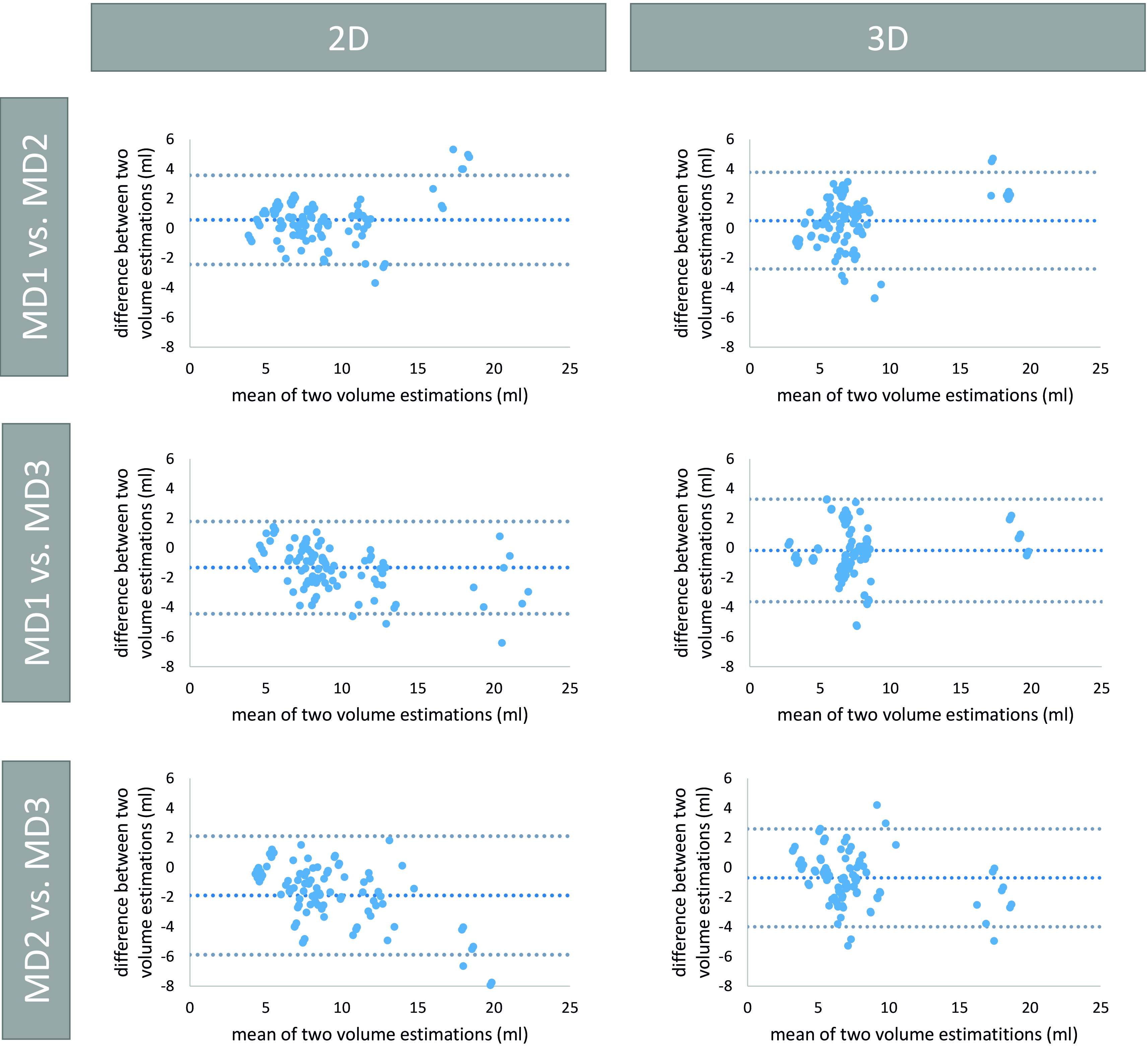}
	\caption{Bland-Altman plots for the interobserver variability between sets of two medical doctors from Group 2. All values are given in $ml$. Row 1, 2 and 3 show results between MD 1 and MD 2, MD 1 and MD 3, and MD 2 and MD 3, respectively. The columns show the results for 2D (left) and 3D measurements (right).}
	\label{fig:7}
\end{figure}

The interobserver variability can be seen in Table~\ref{tab:2} for both groups. The Bland-Altman plots regarding the interobserver variability for Group 2 are depicted in Fig.~\ref{fig:7}.

\begin{table}[h]
	\centering
	\begin{tabular}{c c c c c}
		\hline
		MDs & \multicolumn{2}{c}{2D US} & \multicolumn{2}{c}{p-value}\\
		& Gr. 1 & Gr. 2 & Gr. 1 & Gr. 2\\
		\hline
		1/2 & $1.00\pm1.35$ & $0.58\pm1.53$ & \boldmath $.005$ & $.140$\\
		1/3 & $-2.49\pm1.62$ & $-1.33\pm1.59$ & \boldmath $<.001$ & \boldmath $.002$\\
		2/3 & $-3.49\pm1.43$ & $-1.89\pm2.04$ & \boldmath $<.001$ & \boldmath $.002$\\
		\hline
		MDs & \multicolumn{2}{c}{3D US} & \multicolumn{2}{c}{p-value}\\
		\hline
		1/2 & $0.16\pm1.40$ & $0.52\pm1.62$ & $.582$ & $.176$ \\
		1/3 & $-0.54\pm1.03$ & $-0.17\pm1.03$ & \boldmath $.017$ & $.722$ \\
		2/3 & $-0.70\pm1.15$ & $-0.70\pm1.68$ & \boldmath $.007$ & $.057$ \\
		\hline
	\end{tabular}
	\caption{Mean and standard deviation (mean$\pm$SD) in milliliters ($ml$) of the interobserver variability of two MDs in 2D and 3D for both groups. Significant different p-values are marked in \textbf{bold}.}
	\label{tab:2}
\end{table}

\subsection{Volumetry compared to MRI data}

For the US volume, the first scan series per MD was chosen and compared to the MRI volume. Table~\ref{tab:3} summarizes these comparisons. The mean MRI volume over all volunteers of Group 1 was $7.73 ml$ ($SD=2.71$), while for Group 2 it was $7.00 ml$ ($SD=3.48$).

\begin{table}[h]
	\centering
	\begin{tabular}{c c c c c}
		\hline
		MD & \multicolumn{2}{c}{2D US} & \multicolumn{2}{c}{p-value}\\
		& Gr. 1 & Gr. 2 & Gr. 1 & Gr. 2\\
		\hline
		1 & $8.53\pm2.40$ & $8.38\pm3.26$ & \boldmath $.048$ & \boldmath $.002$\\
		2 & $7.66\pm2.12$ & $8.19\pm3.52$ & $.880$ & \boldmath $.009$\\
		3 & $11.80\pm2.41$ & $10.09\pm4.32$ & \boldmath $<.001$ & \boldmath $<.001$\\
		\hline
		MD & \multicolumn{2}{c}{3D US} & \multicolumn{2}{c}{p-value}\\
		\hline
		1 & $9.09\pm1.80$ & $7.62\pm3.88$ & \boldmath $.018$ & $.292$ \\
		2 & $8.82\pm2.35$ & $6.70\pm3.43$ & \boldmath $.029$ & $.686$ \\
		3 & $9.41\pm2.26$ & $7.55\pm3.57$ & \boldmath $.003$ & $.091$ \\
		\hline
	\end{tabular}
	\caption{Mean and standard deviation (mean$\pm$SD) of the volume of the first scan series for each MD in 2D and 3D in milliliters ($ml$). Statistical significance with respect to the MRI volume is reported, being statistical significant p-values in \textbf{bold}.}
	\label{tab:3}
\end{table}

\subsection{Time}

Time needed for the 2D scans was on average $58.8 s$ ($SD=11.05 s$) for MD1, $55.0 s$ ($SD=8.8 s$) for MD2 and for $35.7 s$ ($SD=5.8 s$) MD3, while the 3D scans were performed on average in $26.0 s$ ($SD=5.5 s$) by MD1, $21.1 s$ ($SD=5.9 s$) by MD2 and in $21.6 s$ ($SD=5.2 s$) by MD3.

\section{Discussion}

\subsection{Comparison with state-of-the-art}

Intra- and interobserver variability in thyroid volumetry is a fairly well analysed problem. Vulpoi et al. had three MDs examine 30 children with 2D US. One physician scanned 25 patients twice to gather data on intraobserver variability, yielding $6.29\%$ ($SD=6.12\%$). The interobserver difference was $9.51\%$ ($SD=8.8\%$)~\cite{vulpoi_thyroid_2007}. Lee et al. studied intraobserver and interobserver variability between doctors with different experience levels (10 years and 6 months) on 122 nodules in 73 volunteers. The two MDs had a similar intraobserver variability $–11.4\%$ ($SD=10.5\%$) and $11.3\%$ ($SD=8.5\%$). The interobserver variability amounted to $15.3\%$ ($SD=16.6\%$)~\cite{lee_intraobserver_2018}. In the study of Choi et al. the thyroid nodule volume of 73 patients with 85 nodules was measured by two MDs with different levels of experience twice. The thyroid nodules were divided into two groups: nodules with a diameter smaller than $2 cm$ and those with a diameter bigger than $2 cm$. In the first group the interobserver variability was $7.0\%$ ($SD=4.9\%$) and $–5.1\%$ ($SD=3.6\%$) in the second group~\cite{lee_intraobserver_2018,choi_inter-observer_2015}.
 
Our study focused on the total thyroid volume. In comparison to the results reported by Vulpoi et al., Lee et al. and Choi et al., our results for intra- and interobserver variability in 2D US (MD1 $14\%$, MD2 $13\%$, MD3 $19\%$ and MD1-MD2 $6.5\%$, MD1-MD3 $-13.12\%$, MD2-MD3 $-19.31\%$) are slightly higher. 

Lyshchik et al. conducted two studies on comparing the volume of thyroid nodules in children both with 2D and 3D US. In the first study one physician scanned 102 children with 129 nodules three times, and delineated their thyroids. The delineated thyroids were controlled by a second expert. The second study included 47 children prior to thyroidectomy. Here, the thyroid weights after surgery were used as ground truth. The nodule volumes were calculated with the ellipsoid formula for 2D and multiplanar volume approximation, respectively manual planimetry, for the 3D scans. Both studies showed very similar results. 3D US showed a lower user dependency than 2D US, had a more accurate thyroid nodule volumetry, was less dependent on nodule size and performed better on irregular nodule outlines. In the first study the intraobserver variability was $16.1\%$ ($SD=0.7\%$) for 2D US and $5.9\%$ ($SD=0.3\%$) for 3D US. The accuracy for 2D US was $15.9\%$ and $2.8\%$ for 3D US. In the second study the intraobserver variability amounted to $14.4\%$ ($SD=1.9\%$) in 2D US and to $3.36\%$ ($SD=0.25\%$) for 3D US. The accuracy amounted to $15.3\%$ for 2D US and $5.2\%$ for 3D US. The second study showed no significant variation between the thyroid weight and both US modalities~\cite{lyshchik_three-dimensional_2004,lyshchik_accuracy_2004}. One limitation of both their studies is that only one physician acquired the data. In our study intraobserver variability for 2D US yielded similar results to Lyshchik et al. Our results for 3D US however are worse. We speculate that the reason for this could be either (a) our network performance in segmenting the thyroid lobes, (b) breathing or movement distorsions as we are using tracked ultrasound instead of a two-dimensional array ultrasound or (c) tracking errors.

With respect to MRI, Reinartz et al. compared 2D US thyroid volumetry to MRI (T1 FFE). Three different experienced MDs performed US scans on patients prior to RIT. The thyroid volumes were calculated by using the ellipsoid formula for the 2D US measurements and by manual segmentation. A significant difference of $22.7\%$ ($10.4 ml$), between the MRI-ROI and mean 2D US measurements was found. No significant interobserver variability among the MDs could be determined~\cite{reinartz_thyroid_2002}. This finding is in contradiction with our results. In our case the MRI volumes were on average $21.3\%$ smaller than the ones obtained from 2D US and $4.99\%$ smaller than the ones obtained from 3D US. This may have to do with the fact that different MRI sequences, resolutions or sampling were used, but most likely by the criteria used for segmentation~\cite{freesmeyer_multimodal_2014}. One study comparing 3D US to CT by Licht et al showed no relevant differences~\cite{licht_3d_2014}.
Rogers et al. analyzed length, diameters and volume measurements of four arteries from a pig using 3D tUS, B-Mode US, CT and MRI. As validation, water immersion technique was used. B-Mode US had the largest error in volume estimation, tUS the smallest. The mean error in volume estimation was $-0.54\pm0,62 ml$ for B-Mode, $-0.06\pm0.09 ml$ for tUS, $0.01\pm0.18 ml$ for CT and $-0.20\pm0.32 ml$ for MRI. These results align with our results in showing that tracked 3D US shows a smaller difference to MRI than 2D US. Furthermore, it shows the second largest mean volume error in volume estimation for MRI. This aligns with our observation of a probable continuous underestimation of our MRI segmentations.

\subsection{Limitations of this study}

Only young and healthy volunteers were enrolled, simplifying the segmentation and volumetry tasks.The CNN was trained with 16 volunteers only, allowing for a good dice score but leaving space for improvements. These can include cross-validation as well as training on 3D input data instead of 2D images. The latter one might allow for additional spatial information which could improve the network performance; yet, it would require a significantly larger dataset. Visual analysis also showed non-optimal segmentation results in few scans (Fig.~\ref{fig:5}). The network seems to be sensitive to the echogenicity of the US images, resulting in additional areas outside of the thyroid to be segmented and areas inside the thyroid to be not segmented. Furthermore, the network was trained on single lobes. This can introduce a potential deviation in the volume due to overlapping or non-segmented parts of the isthmus.
It also has to be noted that not all MDs have used the 3D system before. A longer training could therefore increase the scanning quality.

\section{Conclusion}

Our study shows that 3D US outperforms 2D US significantly. This was consequently seen in both groups 1 and 2 denoting “training” and “new” data. Regarding the intraobserver variability, 3D US outperforms 2D US with 2 out of 3 MDs. In the interobserver variability 3D US increases the similarity of acquisition scans between two MDs in 2 out of 3 cases. The results also show that the improvement with 3D US is the biggest for the least experienced MD. We show that the 3D US volumetry is more accurate than 2D US using MRI as ground truth. The tracked 3D US acquisition is significantly faster as well. Therefore, the combination of tracked 3D US and automatic segmentation is not only feasible but offers a more accurate alternative to conventional US imaging. 
The labelled thyroid 3D US dataset will be freely available helping further projects in deep learning for thyroid volumetry within the research community. 
We suggest  conducting similar studies on patients with thyroid morbidities to stratify the clinical benefit of our framework.

\section*{Acknowledgements}

This work was partially supported by the EU grant 688279 (EDEN2020). The Nuclear Medicine department at KRI thanks Piur Imaging GmbH (www.piurimaging.com; piur imaging GmbH, Hamburgerstr. 11/7, 1050 Vienna, Austria) for a free loan of a PIUR tUS system.

%\section*{Ethical Statement}

%The authors are accountable for all aspects of the work in ensuring that questions related to the accuracy or integrity of any part of the work are appropriately investigated and resolved. The trial was conducted in accordance with the Declaration of Helsinki (as revised in 2013). The study was approved by the Ethical Commission of the Technical University of Munich (approved on April 1st, 2020; reference number 244/19 S) and informed consent was taken from all individual participants.

\appendix


%% References


\bibliographystyle{model1-num-names}
\bibliography{3DThyroid.bib}






% \end{thebibliography}

\end{document}